\documentclass[journal]{IEEEtran}
\usepackage{times}
\usepackage{epsfig}
\usepackage{graphicx}
\usepackage{amsmath}
\usepackage{amssymb}
\usepackage{multirow}
\usepackage{subfigure}
\usepackage{cite}

\ifCLASSINFOpdf
\else
\fi
\begin{document}
%
\title{PcmNet: Position-Sensitive Context Modeling Network for Temporal Action Localization}
%
%
%


\author{Xin~Qin,
	Hanbin~Zhao,
	Guangchen~Lin,
	Hao~Zeng,
	Songcen Xu,
	Xi~Li*
\thanks{X.~Qin, H.~Zhao, G.~Lin, and H.~Zeng are with College of Computer Science, Zhejiang University, Hangzhou 310027, China. (e-mail: xinqin, zhaohanbin, aaronlin, zenghao\_97@zju.edu.cn).}
\thanks{S.~Xu is with Noah's Ark Lab, Huawei Technologies, Shenzhen, 518129, China. (e-mail: xusongcen@huawei.com).}
\thanks{X. Li* (corresponding author) is with the College of Computer Science
		and Technology, Zhejiang University, Hangzhou 310027, China, and also
		with the Alibaba-Zhejiang University Joint Insititute of Frontier Technologies,
		Hangzhou 310027, China (e-mail: xilizju@zju.edu.cn).}
\thanks{The first two authors contributed equally.}
}

%
%

\markboth{IEEE TRANSACTIONS ON IMAGE PROCESSING}%
{Shell \MakeLowercase{\textit{et al.}}: Bare Demo of IEEEtran.cls for IEEE Journals}
%



\maketitle 

\begin{abstract}
Temporal action localization is an important and challenging task that aims to locate temporal regions in real-world untrimmed videos where actions occur and recognize their classes. It is widely acknowledged that video context is a critical cue for video understanding, and exploiting the context has become an important strategy to boost localization performance. However, previous state-of-the-art methods focus more on exploring semantic context which captures the feature similarity among frames or proposals, and neglect positional context which is vital for temporal localization. In this paper, we propose a temporal-position-sensitive context modeling approach to incorporate both positional and semantic information for more precise action localization. Specifically, we first augment feature representations with directed temporal positional encoding, and then conduct attention-based information propagation, in both frame-level and proposal-level. Consequently, the generated feature representations are significantly empowered with the discriminative capability of encoding the position-aware context information, and thus benefit boundary detection and proposal evaluation. We achieve state-of-the-art performance on both two challenging datasets, THUMOS-14 and ActivityNet-1.3, demonstrating the effectiveness and generalization ability of our method.

\end{abstract}


%
\IEEEpeerreviewmaketitle

\section{Introduction}

As an important and challenging problem in human action understanding, temporal action localization (also called temporal action detection) for videos has gained much attention from both academia and industry~\cite{DBLP:conf/eccv/LinZSWY18,DBLP:conf/cvpr/Liu0Z0C19,DBLP:conf/iccv/LinLLDW19,DBLP:conf/cvpr/XuZRTG20,zhao2020bottom,8267510,8105826}, and thus has a large body of applications such as smart surveillance, video retrieval, video abstract, etc.
Temporal action localization aims at detecting the boundaries (\emph{i.e.}, start and end frames) of action instances and predicting their corresponding action classes.
In the literature, the typical pipeline of many existing approaches~\cite{DBLP:conf/eccv/LinZSWY18,DBLP:conf/cvpr/Liu0Z0C19,DBLP:conf/iccv/LinLLDW19,zhao2020bottom} is to detect the start and end boundary frames separately, adaptively generate various action proposals by matching these start and end boundaries, then perform proposal reliability evaluation and temporal boundary regression. Finally, an additional action recognition network is employed to classify each proposal.
Therefore, the quality of temporal action localization mainly depends on precise boundary generation and reliable proposal evaluation.

\begin{figure}[t]
	\begin{center}
		\includegraphics[width=1\linewidth]{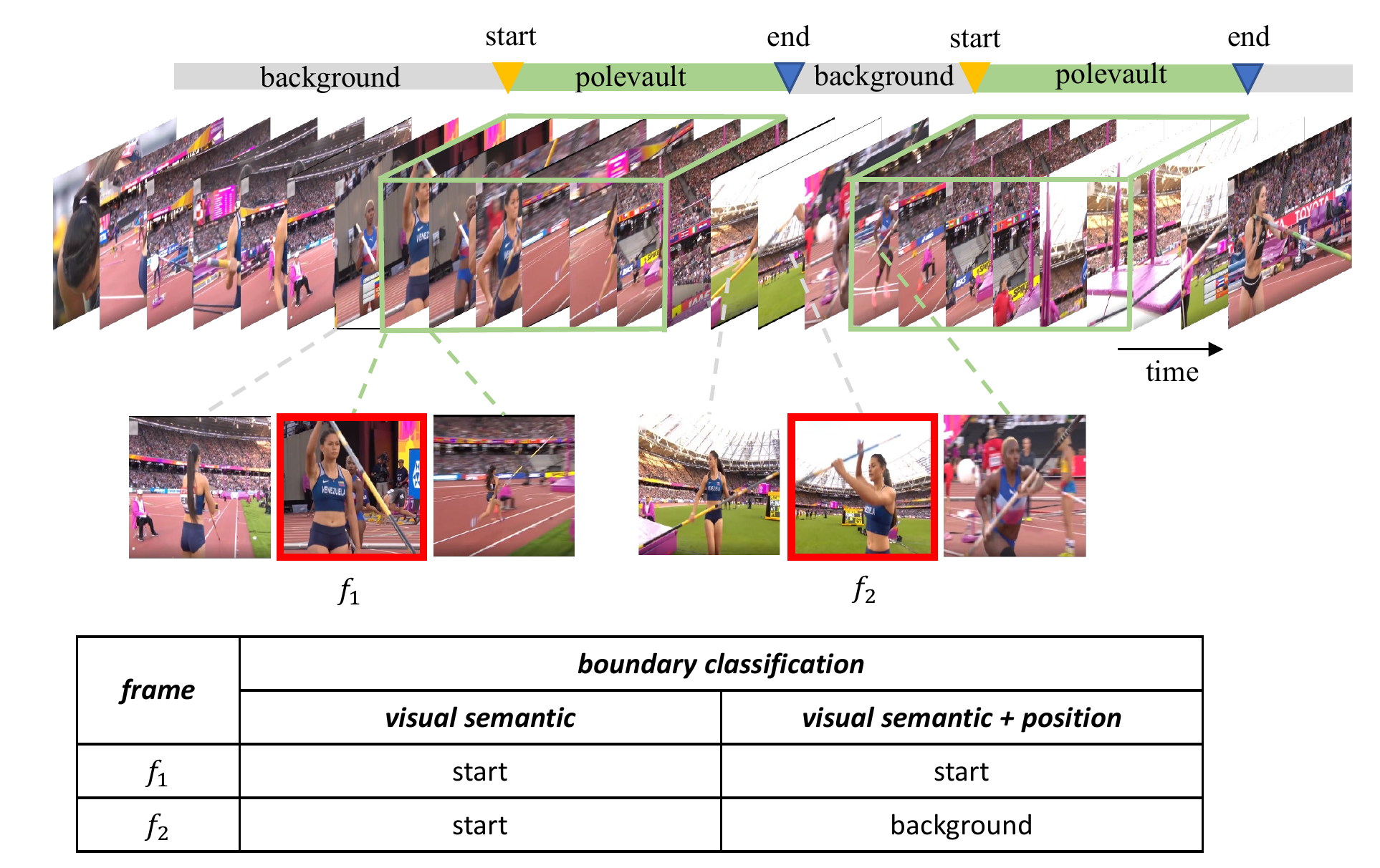}
	\end{center}
	\caption{An example to illustrate the importance of temporal positional context. If only based on the visual semantic content, both the frame $f_1$ and $f_2$ tend to be classified as the same class (\emph{i.e.}, the start frame of the action ``polevault") because of the similar visual semantic content. If based on both the visual semantic content and positional context, these two frames tend to be classified as different classes ($f_1$ is the start frame of action ``polevault", $f_2$ is background). }
	\label{fig:importance_of_location}
\end{figure}

However, the independent features of frames or proposals without context information can be inadequate in determining boundaries or proposal reliability evaluation.
To solve this problem, a number of approaches~\cite{DBLP:conf/iccv/ZhaoXWWTL17,DBLP:conf/cvpr/XuZRTG20,DBLP:conf/cvpr/LongYQTLM19, DBLP:conf/iccv/ZengHGTRZH19} emerge to automatically aggregate the context information among frames (or proposals) for enhancing the feature representation capability.
Such context information is usually characterized by the similarity among frames (or proposals) at the semantic feature level, without modeling the temporal-position context interactions among frames (or proposals) which is valuable prior knowledge. 
As a result, these approaches may suffer from a dilemma that the semantically similar frames can be of different classes at different temporal positions, as shown in Figure~\ref{fig:importance_of_location}.
And the temporal position context for proposals can also specify the regression direction and distance of proposal boundaries to localize actions precisely.


Motivated by these observations, we propose a temporal-position-sensitive context modeling approach, which generates temporal-position-sensitive feature representations by augmenting the semantic features with the directed temporal position encoding information.
Then, through attention-based context propagation in both frame-level and proposal-level, the generated feature representations are significantly empowered with the discriminative capability of encoding the position-sensitive context information.
Specifically, the context of the current frame is associated with two groups of frames (\emph{i.e.}, past frames and future frames), whose position encoding directions are mutually opposite.
And the context of the current proposal corresponds to four groups of proposals generated by imposing four types of coordinate offsets on its start and end positions.
Consequently, the proposed framework takes a progressive context modeling pipeline.
In the pipeline, we first perform frame-level context propagation to produce discriminative frame-level features that help generate more precise action boundaries, and then feed the features into the proposal-level context propagation module to enhance the proposal features for more reliable proposal evaluation. 
In practice, the above two-level context modeling approach is implemented in an attention-based module, and incorporated into a unified learning framework for temporal action localization.

In summary, the main contributions of this work are threefold:
\begin{itemize}
	\item We propose a temporal-position-sensitive context modeling approach, which gives rise to discriminative temporal-position-sensitive feature representations by directed temporal position encoding. To our knowledge, it is innovative and significant to introduce temporal-position-sensitive context modeling into temporal action localization.

	\item We present a unified temporal action localization framework incorporating both frame-level and proposal-level context modeling, which are complementary to each other for more precise temporal action localization.

	\item Extensive experiments on two widely used benchmark datasets demonstrate that our approach outperforms the state-of-the-art method BMN~\cite{DBLP:conf/iccv/LinLLDW19} by $1.4\%$ at average mAP on ActivityNet-1.3~\cite{DBLP:conf/cvpr/HeilbronEGN15}, and $8.4\%$ mAP at IoU@0.5 on THUMOS-14~\cite{THUMOS14}.
\end{itemize}

\begin{figure*}[t]
	\begin{center}
		\includegraphics[width=1\linewidth]{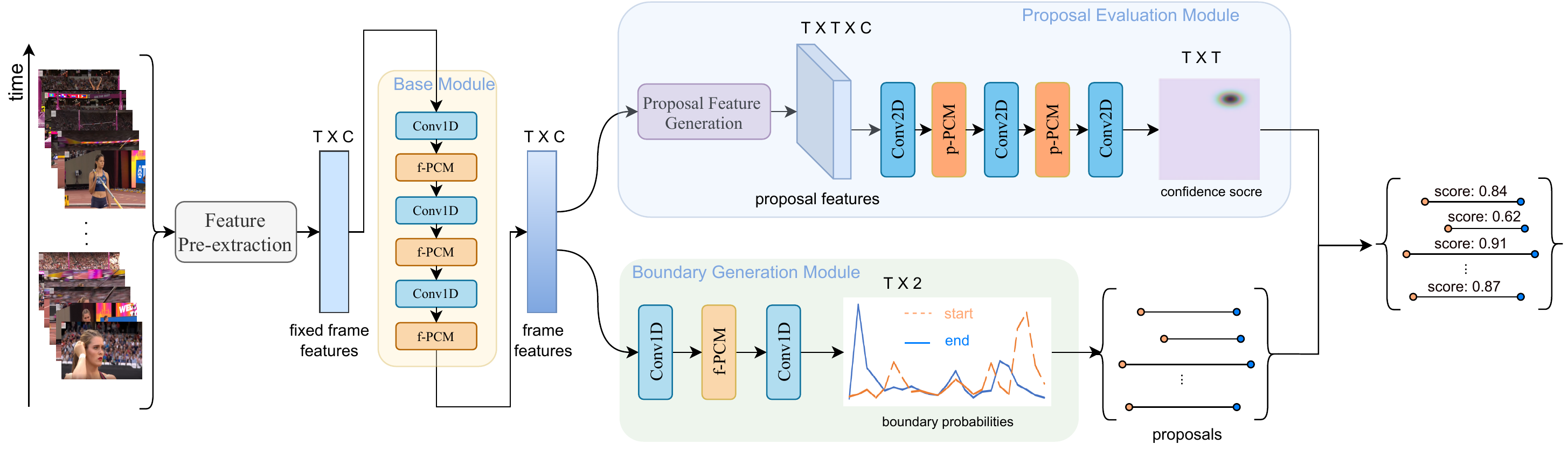}
	\end{center}

	\caption{Overview of our PcmNet. We pre-extract the features from videos and re-scale them to a fixed length. A base module is used to enhance the features. A boundary generation module and a proposal evaluation module are executed in parallel to estimate the boundary probabilities of each frame and evaluate confidence scores for all possible proposals. The proposed temporal-position-sensitive context modeling (PCM) module is employed in the frame-level base module and boundary generation module (f-PCM), as well as proposal-level proposal evaluation module (p-PCM).}

	\label{fig:framework}
\end{figure*}

\section{Related Work}

\subsection{Temporal Action Localization}
Temporal action localization can be decomposed into temporal proposal generation and action classification. A common practice is to first generate temporal proposals and then classify each proposal~\cite{DBLP:conf/eccv/LinZSWY18,DBLP:conf/iccv/LinLLDW19,zhao2020bottom,DBLP:conf/eccv/GaoCN18,DBLP:conf/cvpr/Liu0Z0C19,DBLP:conf/cvpr/XuZRTG20,DBLP:journals/corr/BaiBCGNN20}, because the classification results from the extra action recognition networks~\cite{DBLP:journals/corr/XiongWWZSLL0GT16, DBLP:conf/cvpr/WangXLG17,8600333,8753686,8476540,8360391} are more accurate. There are also some works training these two stages jointly by using a relatively small classification head~\cite{DBLP:conf/iccv/XuDS17,DBLP:conf/cvpr/ChaoVSRDS18} or predicting the locations and classes of actions simultaneously \cite{DBLP:conf/mm/LinZS17,DBLP:conf/bmvc/BuchEGFN17,DBLP:conf/cvpr/YeungRMF16,DBLP:conf/cvpr/LongYQTLM19,DBLP:conf/aaai/LiuW20}. In this paper, we focus on proposal generation and use extra networks to classify the proposals. For proposal generation, there are three main kinds of methods.

\paragraph{Anchor-Based Proposal Generation}
Earlier works~\cite{DBLP:conf/mm/LinZS17,DBLP:conf/cvpr/HeilbronNG16,DBLP:conf/eccv/EscorciaHNG16,DBLP:conf/iccv/GaoYSCN17,DBLP:conf/cvpr/ShouWC16} directly generate proposals with predefined sliding windows or anchors and train a binary classifier for anchor filtering. In order to capture global contextual information for anchors, RapNet~\cite{DBLP:conf/aaai/GaoSWLYGZ20} proposes a relation-aware module to exploit long-range context information among frames, and TALNet~\cite{DBLP:conf/cvpr/ChaoVSRDS18} uses dilation convolution to get a larger receptive field. Although multi-scale anchors~\cite{DBLP:conf/cvpr/ChaoVSRDS18,DBLP:conf/mm/LinZS17} and pyramid architectures~\cite{DBLP:conf/aaai/GaoSWLYGZ20} are used to increase the diversity of anchors, proposals generated by these methods are still not flexible enough to cover actions of varying duration.

\paragraph{Dynamic Proposal Generation}
A more popular pipeline is adaptively forming proposals. On the one hand, some works~\cite{DBLP:conf/eccv/LinZSWY18,DBLP:conf/iccv/ZhaoXWWTL17,DBLP:conf/cvpr/LongYQTLM19, 8846726,8941024,8852682,8897018,8852682} generate proposals by utilizing the continuity of action. They adopt the watershed algorithm to group contiguous frames with high actionness scores as proposals~\cite{DBLP:conf/iccv/ZhaoXWWTL17,8846726}, or predict a Gaussian kernel whose standard deviation represents the interval of a proposal, to represent a proposal directly at each frame and guide context aggregation for proposal feature generation~\cite{DBLP:conf/cvpr/LongYQTLM19}.
On the other hand, some works~\cite{DBLP:conf/eccv/LinZSWY18,DBLP:conf/iccv/LinLLDW19,zhao2020bottom,8941024,8852682,8897018,8852682} generate proposals by boundary matching. They first predict the start and end probabilities for each frame, then match frames with high start and end probabilities.
Graph Convolutional Network (GCN) models are utilized to adaptively incorporate context information on frame level for boundary detection and proposal feature generation~\cite{DBLP:conf/cvpr/XuZRTG20,DBLP:journals/corr/BaiBCGNN20}, or refine the boundaries and classification scores of proposals based on the context among proposals~\cite{DBLP:conf/iccv/ZengHGTRZH19}, by considering frames or proposals as nodes in a graph, respectively.

\paragraph{Complementary Proposal Generation}
Naturally, some works~\cite{DBLP:conf/eccv/GaoCN18,DBLP:conf/cvpr/Liu0Z0C19,9171561,9298475} use anchors to complement the adaptively formed proposal. They collect proposals that could be omitted by actionness grouping from sliding windows~\cite{DBLP:conf/eccv/GaoCN18,9171561}, or adjust the boundaries of the anchors within the search space determined by the computed frame actionness~\cite{DBLP:conf/cvpr/Liu0Z0C19}.

\subsection{Context Modeing in Temporal Action Localization}
Earlier works~\cite{DBLP:conf/cvpr/ChaoVSRDS18,DBLP:conf/iccv/ZhaoXWWTL17,DBLP:conf/eccv/LinZSWY18} simply extend beyond the start and end frames of proposals by half of the proposals duration to take the temporal context into account for proposal classification and location regression. Recently, more and more researchers~\cite{DBLP:conf/iccv/DaiSZDC17,DBLP:conf/iccv/GaoYSCN17,DBLP:conf/iccv/ZhaoXWWTL17,DBLP:conf/cvpr/LongYQTLM19,DBLP:conf/cvpr/XuZRTG20,DBLP:conf/aaai/GaoSWLYGZ20,DBLP:journals/corr/BaiBCGNN20,8933113} have noticed that video context is an important cue to temporal action localization. And they have tried to exploit context to build stronger video representations from different aspects (\emph{e.g.}, frame context~\cite{DBLP:conf/cvpr/LongYQTLM19,DBLP:conf/cvpr/XuZRTG20,DBLP:conf/aaai/GaoSWLYGZ20,8897018}, proposal context~\cite{DBLP:conf/iccv/ZengHGTRZH19,8933113}, boundary-proposal context~\cite{DBLP:journals/corr/BaiBCGNN20}) with various methods (\emph{e.g.}, attention mechanism~\cite{DBLP:conf/aaai/GaoSWLYGZ20,8933113} and GCN~\cite{DBLP:conf/cvpr/XuZRTG20,DBLP:conf/iccv/ZengHGTRZH19,DBLP:journals/corr/BaiBCGNN20}). 
But they focus more on exploring semantic context which aggregates information guided by the feature similarity among frames or proposals. In this paper, we introduce the temporal positional context into temporal action localization, and pay more attention to augment the semantic features with positional information for temporal-position-sensitive context modeling.

\section{Method}

\subsection{Temporal Action Localization}
\label{sec3.1}
We denote the temporal annotation of a video sequence $V$ as $\Psi_g=\{\psi_n=(v^s_{n}, v^e_n,c_n)\}_{n=1}^{N_g}$, where $N_g$ is the number of ground-truth actions, $v^s_{n}$, $v^e_{n}$ and $c_n$ are the start frame, end frame and action class of action $\psi_n$, respectively. The target of temporal action localization is to generate a set of action instances $\Psi_p$ to cover $\Psi_g$ precisely.
Recent works \cite{DBLP:conf/eccv/LinZSWY18,DBLP:conf/iccv/LinLLDW19,zhao2020bottom,DBLP:conf/cvpr/XuZRTG20,DBLP:journals/corr/BaiBCGNN20} pre-extract frame features from raw videos every $\delta$ consecutive frames, with each frame encoded into a vector of $C$ dimension. Then the temporal dimension of the features is rescaled to a fixed length $T$. We then use $\mathbf{X}\in \mathbb{R}^{T \times C}$ to represent the features of the video sequence $V$. 
For simplicity, the input video sequence $V$ is considered as only containing $T$ frames and we denote it as $V=\{v_i\}_{i=1}^{T}$, where $\mathbf{x}_i$ is the feature vector of frame $v_i$.

The temporal action localization is typically conducted by the framework shown in Figure~\ref{fig:framework}. 
The framework mainly consists of three modules: 1) a base module used to enhance the pre-extracted frame features for the next two modules; 2) a boundary generation module used to estimate the boundary probabilities of each frame; 3) a proposal evaluation module to evaluate confidence score of each proposal. These three modules often simply consist of several convolutional layers.

To generate precise proposals, the key issue is to enhance the feature representation ability of these three modules for precise boundary detection and reliable proposal evaluation. A number of approaches emerge to automatically aggregate the context information among frames or proposals for enhancing the feature representation ability, such as utilizing a graph convolutional network (GCN) or an attention block as the boundary generation module~\cite{DBLP:conf/cvpr/XuZRTG20} or proposal evaluation module~\cite{DBLP:conf/iccv/ZengHGTRZH19}. In this paper, we utilize the attention block to aggregate the global context information.

Given the frame features $\mathbf{X}\in \mathbb{R}^{T \times C}$, the attention block aggregates the features of other frames \{$\mathbf{x}_j$\} to generate a new feature $\mathbf{y}_i$ for each frame $v_i$:
\begin{equation}
\mathbf{y}_i =  \frac{1}{\mathcal{C}(\mathbf{X})} \sum_{\forall j} f(\mathbf{x}_i, \mathbf{x}_j)g(\mathbf{x}_j),
\label{vanilla_attention}
\end{equation}
where $f(\cdot,\cdot)$ is a compatibility function to compute the attention weight between the frame $v_i$ and the other frames, $g(\cdot)$ is an unary function to compute the embedding of $\mathbf{x}_j$, $\mathcal{C}(\mathbf{X})=\sum_{\forall j} f(\mathbf{x}_i, \mathbf{x}_j)$ is a normalized factor. 


The attention block can enhance the features of video frames by aggregating the context information. However, it only utilizes the semantic context information, without taking the temporal position information into account. As shown in Figure~\ref{fig:importance_of_location}, such position information is valuable prior knowledge for temporal action localization. To make full use of context in videos, we take both the temporal position and semantic context into account to form a temporal-position-sensitive context modeling (PCM) module. The PCM module is flexible and can be employed in the base module and boundary generation module in frame-level (f-PCM in Section~\ref{section_two_two}), and proposal evaluation module in proposal-level (p-PCM in Section~\ref{section_two_three}). The illustration of our position-sensitive context modeling network (PcmNet) is shown in Figure~\ref{fig:framework}.

\begin{figure}[t]
	\begin{center}
		\includegraphics[width=1\linewidth]{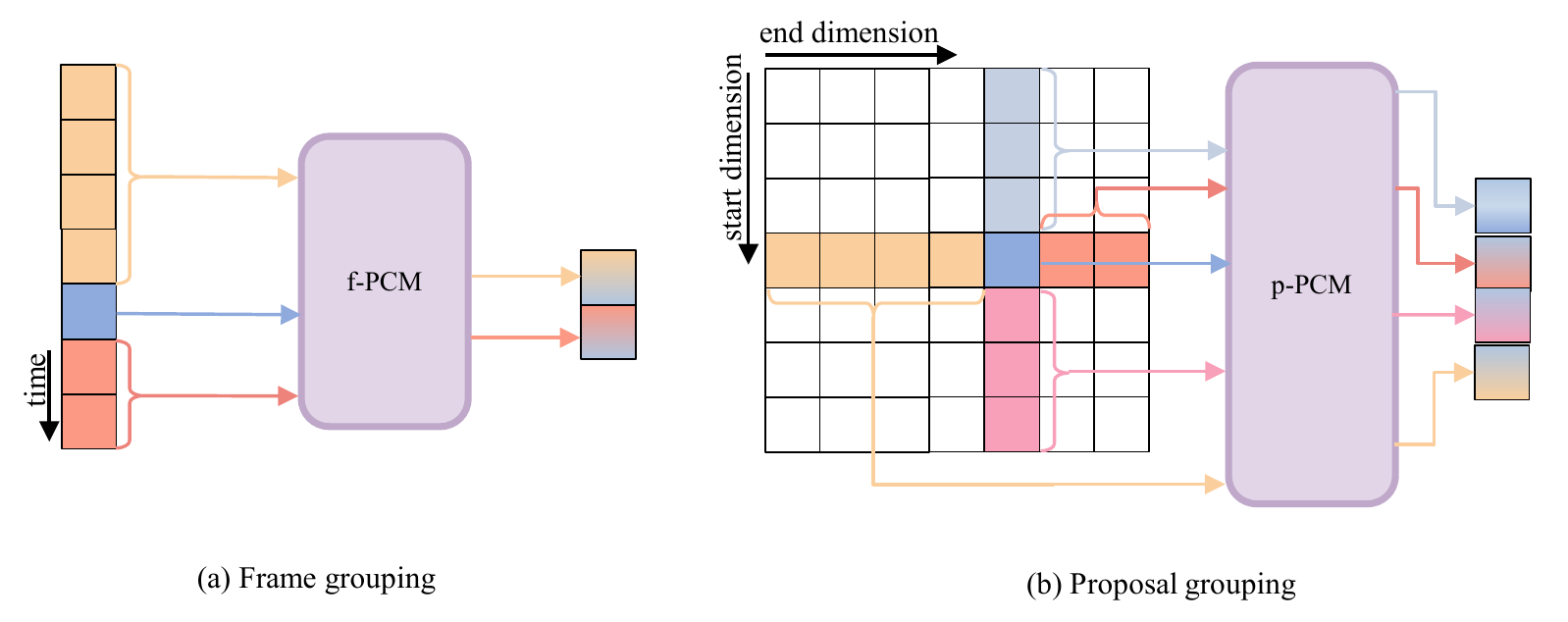}
	\end{center}

	\caption{The illustration of grouping operations in: (a) frame-level and (b) proposal-level. The current frame (proposal) aggregates the context information in each group individually and then orderly concatenates the output features from different groups.}

	\label{fig:PCM}
\end{figure}

\begin{figure*}[t]
	\begin{center}
		\includegraphics[width=0.75\linewidth]{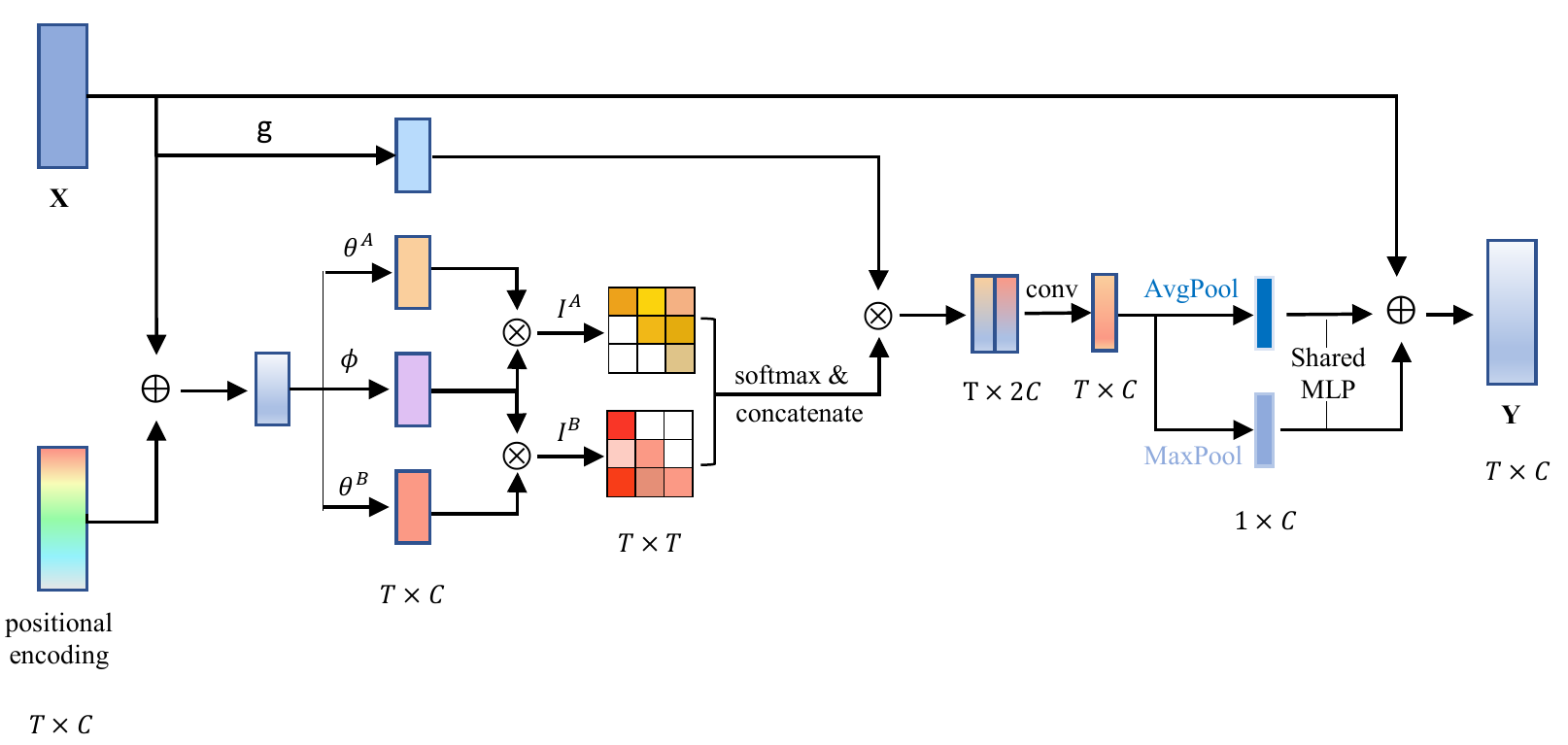}
	\end{center}
	
	\caption{The implementations of the f-PCM. $\otimes$ and $\oplus$ represent Kronecker product and element-wise addition, respectively. }
	
	\label{fig:frame_relation_module}
\end{figure*}

\subsection{Position-Sensitive Frame Context Modeling}\label{section_two_two}

Previous context modeling methods are position-independent, degenerating the sequential frames into an unordered frame-set. However, temporal information is essential to detect boundaries.
To achieve precise boundary detection, the proposed temporal-position-sensitive context modeling module is used in the base module and boundary generation module to preserver the sequential structure when aggregating semantic features.	 

Given the frame features of $\mathbf{X}\in \mathbb{R}^{T \times C}$, our f-PCM generate a new feature for each frame $v_i$ by the following three steps.
Firstly, we generate the position-sensitive frame features $\tilde{\mathbf{X}}$ by adding the positional encoding~\cite{DBLP:conf/nips/VaswaniSPUJGKP17} to $\mathbf{X}$. However, the fixed sinusoid positional encoding can only offer the relative distance information without the direction\footnote{The proof is in the appendix.}. So, secondly, for frame $v_i$, we divide the other frames into two groups according to different relative temporal positions to distinguish the directions. One group contains the frames $V^B=\{v_k\}_{k=1}^{i-1}$ (located before the current frame) and the other one contains the frames $V^A=\{v_k\}_{k=i+1}^{T}$ (located after the current frame). A scheme of the grouping operation is shown in Figure~\ref{fig:PCM}(a).
Finally, we aggregate the features of frames by an attention operation. According to Eq.(\ref{vanilla_attention}), we generate a feature $\mathbf{y}_i^B$ for $v_i$ based on the first group and $\mathbf{y}_i^A$ on the second group with different compatibility function $f^A(\cdot)$ and $f^B(\cdot)$:
\begin{equation}
\mathbf{y}_i^B  =  \frac{1}{\mathcal{C}(\mathbf{X}^B)} \sum_{\forall j, j<i} f^B(\tilde{\mathbf{x}}_i,\tilde{\mathbf{x}}_j)g(\mathbf{x}_j),
\label{eq:y_b}
\end{equation}

\begin{equation}
\mathbf{y}_i^A  =  \frac{1}{\mathcal{C}(\mathbf{X}^A)} \sum_{\forall j, j>i} f^A(\tilde{\mathbf{x}}_i,\tilde{\mathbf{x}}_j)g(\mathbf{x}_j),
\label{eq:y_a}
\end{equation}
In each group, the position-sensitive frame features $\tilde{\mathbf{X}}$ provide temporal distance information when computing the attention weight among the current frame and the other frames. The information in different directions is contained in two groups separately. 
By concatenating the $\mathbf{y}_i^B$ and $\mathbf{y}_i^A$, we can obtain the new feature $\mathbf{y}_i$ for $v_i$:
\begin{equation}
\mathbf{y}_i = \mathbf{y}_i^B \oplus \mathbf{y}_i^A,
\label{eq:f-pcm}
\end{equation}
where $\oplus$ representations the concatenation operation. 
By using a convolutional layer, we can reduce the channel dimension of $\mathbf{y}_i$ from $2C$ to $C$ to aggregate the information in two groups.

\subsection{Position-Sensitive Proposal Context Modeling}\label{section_two_three}

Since there are a lot of redundant and false-positive proposals that are generated by matching start and end frames, it is vital to evaluate the proposals. To generate reliable confidence scores for proposals, we use the temporal-position-sensitive proposal context modeling module (p-PCM) to explore the context among proposals and learn powerful representations.

Different from frames that have inherent temporal sequential structure, proposals cropped from frame sequence are independent and disorderly. Thus, we first organize the proposals into structured data. Specifically, we arrange all possible proposals into a matrix according to the corresponding temporal position of their start and end frames, namely proposal map $\mathbf{M} \in \mathbb{R}^{T \times T}$. An element $m_{i,j}$ in the matrix $\mathbf{M}$ at coordinate $(i, j)$ is the proposal that starts at frame $v_i$ and ends at frame $v_j$. The proposal map indicates the temporal position through the coordinate of the proposals in the matrix.
Based on the frame features from the base module, we can generate a $C$ dimension feature vector for each proposal. We denote the proposal feature map as $\mathbf{P}\in \mathbb{R}^{T \times T \times C}$. The element $\mathbf{p}_{i,j}$ is the feature of the proposal $m_{i,j}$.

Then our p-PCM generates a new feature for each proposal by the following three steps, which are similar to f-PCM. 
We first add the 2D positional encoding to proposal feature map $\mathbf{P}$ to generate the distance-sensitive proposal features $\widetilde{\mathbf{P}}$.
To reduce computational complexity and superfluous context, for each proposal $m_{i,j}$, we only exploit the context of the proposals with the same start frame or end frame as $m_{i,j}$. Aggregating the features of these related proposals is enough to enhance the feature of $m_{i,j}$.
We divide the proposals into four groups based on temporal position to indicate the temporal directions: 1) $M^{EE}=\{m_{i,k} \}_{k=1}^{j-1}$ (containing the proposals which have the same start frame and \textbf{E}arlier \textbf{E}nd frame); 2) $M^{LE}=\{m_{i,k} \}_{k=j+1}^{T}$ (containing the proposals which have the same start frame and \textbf{L}ater \textbf{E}nd frame), 3) $M^{ES}=\{m_{k,j} \}_{k=1}^{i-1}$ (containing the proposals which have the same end frame and \textbf{E}arlier \textbf{S}tart frame) and 4) $ M^{LS}=\{m_{k,j} \}_{k=i+1}^{T}$ (containing the proposals which have the same end frame and \textbf{L}ater \textbf{S}tart frame). A scheme of the proposal grouping operation is shown in Figure~\ref{fig:PCM}(b).
At last, for each group, we aggregate the features of proposals by an attention operation. According to Eq.(\ref{vanilla_attention}), we generate proposal features $\mathbf{py}_{i,j}^{EE}, \mathbf{py}_{i,j}^{LE}, \mathbf{py}_{i,j}^{ES}$, $\mathbf{py}_{i,j}^{LS}$ based on the four groups separately.
By concatenating these four proposal feature maps, we can obtain the enhanced proposal feature $\mathbf{py}_{i,j}$ for proposal $m_{i,j}$.
A convolutional layer is used to reduce the channel dimension from $4C$ to $C$.

\subsection{Implementation}
The computation of PCM can be simply done by matrix multiplication. We take f-PCM as an example.
We choose the embedded Gaussian function as the compatibility function in Eq.(\ref{eq:y_b}) and (\ref{eq:y_a}):

\begin{equation}
f^B(\tilde{\mathbf{x}}_i,\tilde{\mathbf{x}}_j)= {e ^ {\phi^B (\mathbf{\tilde{x}}_i) \theta^B (\mathbf{\tilde{x}}_j)^\mathsf{T} }},
\end{equation}

\begin{equation}
f^A(\tilde{\mathbf{x}}_i,\tilde{\mathbf{x}}_j)= {e ^ {\phi^A (\mathbf{\tilde{x}}_i) \theta^A (\mathbf{\tilde{x}}_j)^\mathsf{T} }},
\end{equation}
where $\theta^B(\cdot)$ ,$\phi^B(\cdot)$, $\theta^A(\cdot)$ ,$\phi^A(\cdot)$ are the embedding functions. 
These four embedding functions as well as $g(\cdot)$ are all implemented as convolutional layers, the normalization is implemented by a softmax function. As shown in Figure~\ref{fig:frame_relation_module}, given the input frame features $\mathbf{X} \in \mathbb{R}^{T \times C}$ and positional encoding, the outputs $\mathbf{Y}^B$ and $\mathbf{Y}^A$ are obtained as following:
\begin{equation}
\mathbf{Y}^B = \sigma(\phi^B (\mathbf{\tilde{X}})  \theta^B (\mathbf{\tilde{X}})^\mathsf{T}  \mathbf{I}^B)  g(\mathbf{X}),
\label{eq:f-pcm}
\end{equation}
\begin{equation}
\mathbf{Y}^A = \sigma(\phi^A (\mathbf{\tilde{X}})  \theta^A (\mathbf{\tilde{X}})^\mathsf{T}  \mathbf{I}^A)  g(\mathbf{X}),
\label{eq:f-pcm}
\end{equation}
where $\sigma$ is the softmax function, $\mathbf{I}^B$ and $\mathbf{I}^A$ represent the lower triangular mask matrix and upper triangular mask matrix for grouping, $\phi^A(\cdot)$ and $\phi^B(\cdot)$ share the same parameters. In practice, we find that using such complex operations in a shallow neural network carries the potential risk of overfitting. Inspired by the channel attention mechanism \cite{DBLP:conf/cvpr/HuSS18, DBLP:conf/eccv/WooPLK18}, we only modulate the channel weights to alleviate overfitting. 

As for p-PCM, since we only exploit the context of the proposals with the same start frame or end frame as $m_{i,j}$ and they are in the same row or column as $m_{i,j}$ in proposal map $\mathbf{M}$, we can decompose the attention on the proposal feature map into column-only attention followed by row-only attention, where each attention can be implemented similar to f-PCM.

\subsection{Training and Inference}
\paragraph{Training.}
We train our PcmNet in the form of a multi-task loss function, including a boundary detection loss term $L_b$ and a proposal evaluation loss term $L_p$:
\begin{equation}
\begin{aligned}
L = L_b + L_p.
\end{aligned}
\label{eq:loss}
\end{equation}
The loss $L_b$ is used to classify whether each frame is the start or the end boundary of an action:
\begin{equation}
\begin{aligned}
L_b = L_{wce}(\mathbf{pred}^s, \mathbf{gt}^s) + L_{wce}(\mathbf{pred}^e, \mathbf{gt}^e),
\end{aligned}
\label{eq:L_b}
\end{equation}
where $L_{wce}$ is the weighted binary cross-entropy loss function proposed in~\cite{DBLP:conf/eccv/LinZSWY18}, $\mathbf{pred}^s$ are the predicted start probability vector of frames ($\mathbf{pred}^e$ for the predicted end probability), $\mathbf{gt}^s$ and $\mathbf{gt}^e$ are the binary ground-truth.  
The loss $L_p$ is used for proposal confidence prediction. Following~\cite{DBLP:conf/iccv/LinLLDW19}, we predict two confidence map $\mathbf{Mcc}$ (trained by the weighted binary cross-entropy loss function) and $\mathbf{Mcr}$ (trained by the L2 regression loss function):
\begin{equation}
\begin{aligned}
L_p = L_{wce}(\mathbf{Mcc}, \mathbf{IoU}^{GT}) + \lambda L_2(\mathbf{Mcr}, \mathbf{IoU}^{GT}) ,
\end{aligned}
\label{eq:L_p}
\end{equation}
where $\lambda$ is the loss weight, which is set to 10 as default in our experiments.

\paragraph{Inference.}
We would obtain start and end boundary probabilities $\mathbf{pred}^s, \mathbf{pred}^e$, and proposal confidence map $\mathbf{Mcc}$ and $\mathbf{Mcr}$ from the network outputs.
To get final results, we need to go through two more steps: proposal generation and redundant proposal suppression.
Firstly, we select the frames with their start/end probability higher than a threshold or are probability peaks as start/end frames. Then, we match each start frame $v_i$ and end frame $v_j$ as a proposal if $i<j$, and calculate its confidence score $cs_{i,j}={Mcc}_{i,j} \cdot {Mcr}_{i,j}$.   
Thus, we obtain a set of proposals $\Psi_p = \{\psi_n=(v^s_{n}, v^e_{n}, {cs}_n)\}^{N_c}_{n=1}$.
At last, we adopt a Soft-NMS algorithm \cite{DBLP:conf/iccv/BodlaSCD17} to eliminate the redundant proposals and get the final proposals set $\Psi_p = \{\psi_n=(v^s_{n}, v^e_{n}, {cs}_n)\}^{N_p}_{n=1}$, where the number of proposals is decreased from $N_c$ to $N_p$.

\begin{table*}[t]
	\begin{center}
		\caption{Action localization results on the test set of THUMOS-14 and validation set of ActivityNet-1.3, measured by mAP (\%) at different tIoU thresholds (and the average mAP on ActivityNet-1.3). * indicates that we reproduce the results with their released code by replacing their I3D features with 2D TSN features on THUMOS-14 dataset for a fair comparison.}
		\label{tab:sota_results}
		\resizebox{1\textwidth}{!}{
			\begin{tabular}{ccc||ccccc||cccc}
				\hline
				\multirow{2}{*}{Method}  & \multirow{2}{*}{Year}  & \multirow{2}{*}{Feature}
				& \multicolumn{5}{c||}{THUMOS-14} & \multicolumn{4}{c}{ActivityNet-1.3}  \\
				& & & 0.3 & 0.4 & 0.5 & 0.6 & 0.7 & 0.5 & 0.75 & 0.95 & Avg. \\
				
				\hline
				SCNN~\cite{DBLP:conf/cvpr/ShouWC16} 		&CVPR'16 & DTF  		& 36.3 & 28.7 & 19.0 & 10.3 &  5.3 &  -    & -     &  -    & -     \\
				SCC~\cite{DBLP:conf/cvpr/HeilbronBEG17} 	&CVPR'17 & C3D   		& -    & -    & -    & -    & -    & 39.9  & 18.7  &  4.7  & 19.3  \\
				CDC~\cite{DBLP:conf/cvpr/ShouCZMC17}    	&CVPR'17 & C3D   		& 40.1 & 29.4 & 23.3 & 13.1 &  7.9 & 45.3  & 26.0  &  0.2  & 23.8  \\
				TURN~\cite{DBLP:conf/iccv/GaoYSCN17} 		&ICCV'17 & Flow  		& 44.1 & 34.9 & 25.6 &  -   &   -  &   -   &   -   &   -   & -     \\
				TCN~\cite{DBLP:conf/iccv/DaiSZDC17}     	&ICCV'17 & TSN   		& -    & 33.3 & 25.6 & 15.9 & 9.0  & 37.49 & 23.47 & 4.47  & 23.58  \\
				SSN~\cite{DBLP:conf/iccv/ZhaoXWWTL17}		&ICCV'17 & TSN   		& 51.9 & 41.0 & 29.8 & -    & -    & 39.12 & 23.48 & 5.49  & 23.98  \\
				TALNet~\cite{DBLP:conf/cvpr/ChaoVSRDS18} 	&CVPR'18 & I3D   		& 53.2 & 48.5 & 42.8 & 33.8 & 20.8 & 38.23 & 18.30 & 1.30  & 20.22  \\
				BSN~\cite{DBLP:conf/eccv/LinZSWY18} 		&ECCV'18 & TSN  		& 53.5 & 45.0 & 36.9 & 28.4 & 20.0 & 46.45 & 29.96 & 8.02  & 30.03  \\
				MGG~\cite{DBLP:conf/cvpr/Liu0Z0C19} 		&CVPR'19 & TSN   		& 53.9 & 46.8 & 37.4 & 29.5 & 21.3 &   -   &   -   &   -   &   -    \\
				GTAN~\cite{DBLP:conf/cvpr/LongYQTLM19} 		&CVPR'19 & P3D   		& 57.8 & 47.2 & 38.8 & -    &   -  & \textbf{52.61} & 34.14 & 8.91 & 34.31   \\
				PGCN~\cite{DBLP:conf/iccv/ZengHGTRZH19}		&ICCV'19 & TSN*, I3D    & 60.1 & 54.3 & 45.5 & 33.5 & 19.8 & 48.26 & 33.16 & 3.27  & 31.11  \\
				BMN~\cite{DBLP:conf/iccv/LinLLDW19}			&ICCV'19 & TSN 			& 56.0 & 47.4 & 38.8 & 29.7 & 20.5 & 50.07 & 34.78 &  8.29 & 33.85  \\
				SRG~\cite{8897018}							&TCSVT'20& TSN 			& 54.5 & 46.9 & 39.1 & 31.4 & 22.2 & 45.53 & 29.98 &  4.83 & 29.72  \\
				A2Net~\cite{8897018}						&TIP'20  & I3D 			& 58.6 & 54.1 & 45.5 & 32.5 & 17.2 & 43.55 & 28.69 &  3.70 & 27.75  \\
				GTAD~\cite{DBLP:conf/cvpr/XuZRTG20} 		&CVPR'20 & TSN 			& 54.5 & 47.6 & 40.2 & 30.8 & 23.4 & 50.36 & 34.60 &  9.02 & 34.09  \\
				BU-MR~\cite{zhao2020bottom} 				&ECCV'20 & TSN*,I3D     & 52.4 & 48.6 & 42.5 & 35.6 & 27.0 & 43.47 & 33.91 & 9.21  & 30.12  \\
				BC-GNN~\cite{DBLP:journals/corr/BaiBCGNN20} &ECCV'20 & TSN          & 57.1 & 49.1 & 40.4 & 31.2 & 23.1 & 50.56 & 34.75 & 9.37 & 34.26  \\
				
				\hline
				
				\multicolumn{2}{c}{PcmNet (Ours)}	     & TSN  & \textbf{61.5} & \textbf{55.4} & \textbf{47.2} & \textbf{37.5} & \textbf{27.3} & {51.35} & \textbf{36.10} & \textbf{9.49} & \textbf{35.27}   \\
				\hline
			\end{tabular}
		}
	\end{center}
\end{table*}

\section{Experiment}

\subsection{Datasets and Metrics}

\paragraph{THUMOS-14~\cite{THUMOS14}} It contains 413 temporal annotated untrimmed videos with 20 action categories. We use 200 videos in the validation set for training and evaluate on 213 videos in the testing set. This dataset is challenging as some videos are relatively long (up to 26 minutes) and contain multiple actions. The length of actions varies from less than a second to minutes.

\paragraph{ActivityNet-1.3~\cite{DBLP:conf/cvpr/HeilbronEGN15}} It is a large-scale action understanding dataset, which consists of 19994 videos for training, 4728 for validation, and 5044 for testing, with 200 activity classes. The total duration of the videos is about 600 hours. The ActivityNet-1.3 only contains 1.5 occurrences per video on average and most videos simply contain a single action category with 36\% background on average.

\paragraph{Metrics} Mean Average Precision (mAP) is the evaluation metric of temporal action localization task, where Average Precision (AP) reports the performance of each activity category. Following the official evaluation API, mAP with IoU thresholds $\{0.5, 0.75, 0.95\}$ and average mAP with IoU thresholds $[0.5 : 0.05 : 0.95]$ are used on ActivityNet-1.3, while mAP with IoU thresholds $\{0.3, 0.4, 0.5, 0.6, 0.7\}$ are used on THUMOS-14.

\subsection{Implementation Details}
\paragraph{Feature Encoding}
Following previous works~\cite{DBLP:conf/eccv/LinZSWY18,DBLP:conf/iccv/LinLLDW19,DBLP:conf/cvpr/XuZRTG20}, we adopt the 2D two-stream network~\cite{DBLP:conf/eccv/WangXW0LTG16} to pre-extract video features for both datasets. For ActivityNet-1.3, we adopt the two-stream network fine-tuned on the training set of ActivityNet-1.3 with frame interval $\sigma = 16$. Each video feature sequence is rescaled to $T = 100 $ frames by linear interpolation. For THUMOS-14, the video features are extracted from Kinetics~\cite{DBLP:journals/corr/KayCSZHVVGBNSZ17} pre-trained two-stream network with $\sigma = 5$. We crop each video feature sequence with a window size $L = 256$ and overlap neighborhood windows with $128$ frames. As for proposal temporal length alignment and feature generation, we follow the previous work~\cite{DBLP:conf/iccv/LinLLDW19}.

\paragraph{Training and Inference}
The network is trained from scratch, and we set the weight decay to $ 1 \times 10^{-4} $, batch size to 32 for both datasets. The learning rate is $ 1 \times 10^{-3} $ for the first 7 epochs, and is reduced by 10 for the following 3 epochs on ActivityNet-1.3, and $ 7 \times 10^{-5} $ for the first 5 epochs, and reduced by 10 for the following 5 epochs on THUMOS-14. 
In inference, we take video classification scores by \cite{DBLP:conf/cvpr/WangXLG17} for THUMOS14-14 and \cite{DBLP:journals/corr/XiongWWZSLL0GT16} for ActivityNet-1.3, and multiply their classification scores to proposal confidence. For post-processing, the threshold probability for selecting start and end frames is set to be half of the maximum probability in the video, and the Soft-NMS threshold is set to 0.5 to pick the top $N_p$ confident predictions, where $N_p$ is 100 for ActivityNet-1.3 and 200 for THUMOS-14.

\subsection{Comparison with State-of-the-Art}


In Table \ref{tab:sota_results}, we compare our PcmNet with state-of-the-art methods on two widely used datatsets, THUMOS-14 and ActivityNet-1.3, at different tIoU thresholds as well as average mAP on ActivityNet-1.3.
On \textit{THUMOS-14}, we achieve the highest mAP at all tIoU thresholds.
As for \textit{ActivityNet-1.3}, we achieve the highest average mAP on this large-scale and diverse dataset. It is worth noting that we achieve the best results at tIoU 0.75 and 0.95, indicating that we localize actions more accurately.
We outperform BMN which is without context modeling by $8.4\%$ mAP at IoU@0.5 on THUMOS-14~\cite{THUMOS14} and $1.4\%$ at average mAP on ActivityNet-1.3~\cite{DBLP:conf/cvpr/HeilbronEGN15}. 
And compared to other context modeling methods (SRG,A2Net,GTAD,BU-MR,BC-GNN), our PcmNet also surpasses them by a large margin, manifesting the effectiveness of our position-sensitive context modeling.
It should be noted that it is difficult to improve performance on the large-scale ActivityNet-1.3 dataset. As a comparison, the improvements of the latest SOTA BC-GNN~[1]~(ECCV20) and G-TAD~[31]~(CVPR20) are $0.39\%$ and $0.24\%$ respectively compared to BMN~[16]~(ICCV19). The large performance improvements demonstrate the effectiveness of our method.

\begin{table}[t]
	\begin{center}
		\caption{Validation of f-PCM and p-PCM.}
		\label{table:FCM_PCM}
		\resizebox{0.5\textwidth}{!}{
			\begin{tabular}{c|c|ccc|ccc}
				\hline
				\multirow{2}{*}{f-PCM}  & \multirow{2}{*}{p-PCM}
				& \multicolumn{3}{c|}{THUMOS-14} & \multicolumn{3}{c}{ActivityNet-1.3}  \\
				&  & 0.3  & 0.5 &  0.7 & 0.5 & 0.75 & Avg. \\
				\hline
				           & 		   	 & 56.12 & 42.01 & 22.78 & 50.72 & 35.05 & 34.17  \\
				\checkmark & 			 & 60.35 & 45.25 & 26.39 & 50.95 & 35.84 & 34.88  \\
						   & \checkmark  & 60.02 & 44.67 & 25.72 & 51.00 & 36.11 & 34.91  \\
				\checkmark & \checkmark  & \textbf{61.49} & \textbf{47.17} & \textbf{27.28} & \textbf{51.35} & \textbf{36.10} & \textbf{35.27}  \\		
				\hline
			\end{tabular}
		}
	\end{center}

\end{table}

There are two interesting phenomenons. 
One is that many works perform well on one dataset, but not so well on another. For example, TALNet achieves significant improvement on THUMOS-14 in 2018, and outperforms BSN in the same period at tIoU 0.5 by a large margin (5.9\%). However, on ActivityNet-1.3, its average mAP is much worse (9.81\%) than BSN. The same thing happens on BU-MR and PGCN while BC-GNN is the opposite. 
Another is that some methods achieve high performance at small tIoU but deteriorate rapidly as tIoU increases, like PGCN and A2Net.
Notably, we reach the highest level on both datasets and all tIoU, indicating that the proposed method has good generalizability.

\begin{figure*}[t]
	\begin{center}
		\includegraphics[width=1\linewidth]{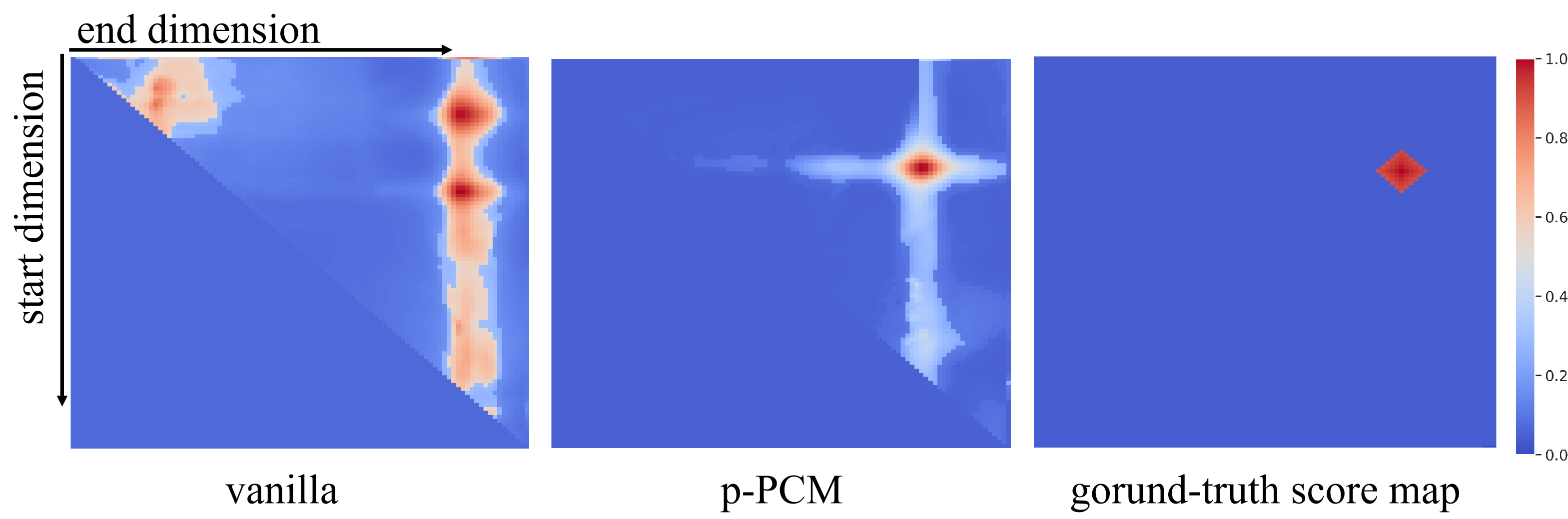}
	\end{center}
	\caption{Visualization for the average attention map of all query proposals. The attention maps of our p-PCM are well aligned to the ground-truth proposal score map, while that of the vanilla attention drift from ground-truth.}
	\label{fig:vis_ppcm2}
\end{figure*}

\subsection{Ablation Study}
We perform ablation studies on both two datasets to validate the effectiveness and generalizability of our method. We first study the effectiveness and complementarity of our position-sensitive context modeling approach in frame-level (f-PCM) and proposal-level (p-PCM). Then, we look into the impact of different components in f-PCM and p-PCM separately. At last, we compare our PCM modules with other context modeling methods in terms of performance and speed to manifest the effectiveness and efficiency, and combine our PCM modules with other frameworks to justify their generalization.

As described in Section~\ref{sec3.1}, we follow the pipeline of BMN with some adjustments of network structure (details can be seen in Fig~\ref{fig:framework}) as the baseline.

\begin{table}[t]
	\begin{center}
		\caption{Validation of different components in f-PCM. SA, CA, GP and PE represent self-attention, channel attention, grouping and positional encoding, respectively.}
		\label{table:FCM}
		\resizebox{0.5\textwidth}{!}{
			\begin{tabular}{l|ccc|ccc}
				\hline
				\multirow{2}{*}{Method}
				& \multicolumn{3}{c|}{THUMOS-14} & \multicolumn{3}{c}{ActivityNet-1.3}  \\
				& 0.3  & 0.5 &  0.7 & 0.5 & 0.75 & Avg. \\
				\hline
				{Baseline} 				& 56.12 & 42.01 & 22.78 & 50.72 & 35.05 & 34.17 \\	
				{Baseline+SA} 			& 58.56 & 44.84 & 23.94 & 49.94 & 34.62 & 33.74  \\
				{Baseline+SA+CA} 		& 57.97 & 44.21 & 23.93 & 50.78 & 35.68 & 34.54  \\
				{Baseline+SA+CA+GP} 	& 60.00 & 45.00 & 25.85 & 50.91 & \textbf{35.86} & 34.82  \\
				{Baseline+SA+CA+PE} 	& 59.15 & 44.57 & 25.59 & 50.90 & 35.84 & 34.78  \\
				{Baseline+SA+CA+GP+PE}  & \textbf{60.35} & \textbf{45.25} & \textbf{26.39} & \textbf{50.95} & 35.84 & \textbf{34.88}  \\			
				\hline
			\end{tabular}
		}
	\end{center}

\end{table}

\subsubsection{Effectiveness and Complementarity}
To study the effectiveness and complementarity of f-PCM and p-PCM, we show the performance improvement of different combinations of them in Table~\ref{table:FCM_PCM}. The f-PCM delivers $3.24\%$ mAP at tIoU 0.5 on THUMOS-14 and $0.71\%$ average mAP on ActivityNet-1.3 performance improvement, and p-PCM delivers $2.66\%$ mAP at tIoU 0.5 on THUMOS-14 and $0.74\%$ average mAP on ActivityNet-1.3. When integrating both f-PCM and p-PCM, it exceeds baseline by $5.16\%$ mAP at tIoU 0.5 on THUMOS-14 and $1.1\%$ average mAP on ActivityNet-1.3. These experiments demonstrate that both frame-level and proposal-level position-sensitive context modeling are helpWful for temporal action localization, and these two-level contexts are complementary with each other for more precise localization.

\begin{table}[t]
	\begin{center}
		\caption{Validation of different components in p-PCM. DSA, CA, GP and PE represent our decomposed self-attention, channel attention, grouping and positional encoding, respectively.}
		\label{table:PCM}
		\resizebox{0.5\textwidth}{!}{
			\begin{tabular}{l|ccc|ccc}
				\hline
				\multirow{2}{*}{Method}
				& \multicolumn{3}{c|}{THUMOS-14} & \multicolumn{3}{c}{ActivityNet-1.3}  \\
				& 0.3  & 0.5 &  0.7 & 0.5 & 0.75 & Avg. \\
				\hline
				{Baseline} 				& 56.12 & 42.01 & 22.78 & 50.72 & 35.05 & 34.17 \\	
				{Baseline+DSA} 			& 57.09 & 43.60 & 24.47 & 50.19 & 35.38 & 34.64   \\
				{Baseline+DSA+CA} 		& 58.06 & 43.90 & 24.13 & 50.82 & 35.54 & 34.69  \\
				{Baseline+DSA+CA+GP} 	& 59.69 & \textbf{44.87}& 25.16 & 50.93 & 35.99 & 34.87  \\
				{Baseline+DSA+CA+PE}    & 59.38 & 44.62 & 25.07 & 50.90 & 35.89 & 34.83  \\
				{Baseline+DSA+CA+GP+PE}  & \textbf{60.02} & 44.67 & \textbf{25.72} & \textbf{51.00} & \textbf{36.11} & \textbf{34.91} \\			
				\hline
			\end{tabular}
		}
	\end{center}
\end{table}

\subsubsection{Frame Context Modeling Module}
To figure out how much of a role each component in f-PCM plays, we carry out detailed verification experiments for different combinations of self-attention, channel attention, grouping and positional encoding in Table~\ref{table:FCM}. Baseline means we run the model without any attention. The self-attention improves performance on THUMOS-14 dataset, but leads to performance degradation on ActivityNet-1.3 dataset. A possible explanation is that the complex self-attention runs the risk of overfitting. Compared to THUMOS-14, the data distribution of ActivityNet-1.3 is more diverse and the validation set has more patterns not seen in the training set, so that the risk of overfitting on this dataset is greater. The channel attention acts as a regularization here, which only modulates the channel weights rather than generating new features. On the THUMOS-14 dataset, channel attention does not provide any performance improvement compared to self-attention. However, grouping and positional encoding both make good performance on two datasets. And it can already achieve reasonable results with grouping operation only, that is, distinguishing the direction provides a strong cue when propagating information. Positional encoding can further bring about slight improvement with negligible computational cost.

\begin{table}[t]
	\begin{center}
		\caption{Comparison of different context modeling methods. We compare our f-PCM with frame-level context modeling methods (\emph{i.e.}, 1D NL, GCN), and our p-PCM with proposal-level context modeling methods (\emph{i.e.}, 2D NL, GCN). ``NL" represents non-local attention.}
		\label{table:different_context_modeling_methods}
		\resizebox{0.5\textwidth}{!}{
			\begin{tabular}{c|c|cc|ccc}
				\hline
				\multirow{2}{*}{Method}
				& \multicolumn{3}{c|}{THUMOS-14} & \multicolumn{3}{c}{ActivityNet-1.3} \\
				& latency (s) & 0.5 &  0.7 & 0.5 & 0.75 & Avg. \\
				\hline
				Baseline   	   	    & 0.0367 & 42.01 & 22.78 & 50.72 & 35.05 & 34.17   \\
				\hline	
				1D NL     			& 0.0407 & 44.84 & 23.94 & 49.94 & 34.62 & 33.74  \\
				GCN    		 		& 0.0613 & 43.60 & 23.47 & 50.78 & 35.09 & 34.25  \\
				f-PCM   		 	& 0.0416 & \textbf{45.25} & \textbf{26.39} & \textbf{50.95} & \textbf{35.84} & \textbf{34.88}   \\
				\hline
				2D NL    			& 0.1117 & 43.32 & 23.96 & \textbf{51.03} & 35.40 & 34.48  \\
				GCN    		 		& 1.7213 & \textbf{45.49} & 19.83 & 48.26 & 33.16 & 31.11   \\
				p-PCM   		 	& 0.0628 & 44.67 & \textbf{25.72} & 51.00 & \textbf{36.11} & \textbf{34.91}    \\	
				\hline
			\end{tabular}
		}
	\end{center}
	
\end{table}

\begin{figure*}[t]
	\begin{center}
		\includegraphics[width=0.85\linewidth]{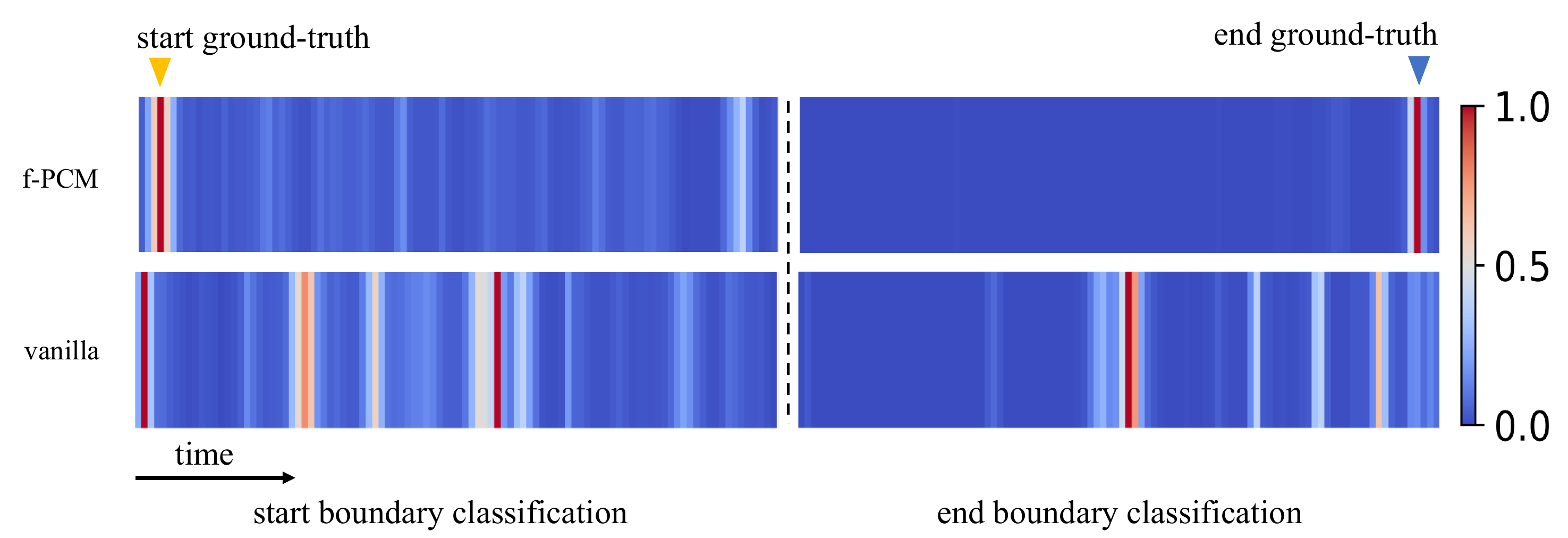}
	\end{center}
	
	\caption{Visualization for the average attention maps before start and end classification (left column is for the start frame and the right is for the end frame). Our f-PCM pays more attention to the true start and end frames while vanilla attention drifts away from the true positions. }
	
	\label{fig:vis_fpcm}
	
\end{figure*}

\subsubsection{Proposal Context Modeling Module}
We then perform similar tests to figure out the roles played by each component in p-PCM. The results are shown in Table~\ref{table:PCM}. Different from f-PCM, the self-attention improves performance on both two datasets and the channel attention operation provides little improvement on the two datasets. Similar to f-PCM, grouping and positional encoding bring improvements in both datasets. The performance improvements bring from self-attention, proposal grouping and positional encoding are $1.59\%$, $0.83\%$ and $0.68\%$ at tIoU 0.5 on THUMOS-14. The self-attention in Table ~\ref{table:PCM} is different from that in Table ~\ref{table:FCM}. The former is decomposed and lightweight based on our proposed context fusion rules the proposal map while the latter is the normal attention. This decomposed self-attention achieve better performance with less computation (the performance of normal attention is shown in Table~\ref{different_framework}).

\subsubsection{Different Context Modeling Methods}
We compare our f-PCM and p-PCM with other context modeling methods, GCN and vanilla attention, to demonstrate the importance of the temporal position information. For GCN, G-TAD~\cite{DBLP:conf/cvpr/XuZRTG20} and PGCN~\cite{DBLP:conf/iccv/ZengHGTRZH19} model the context in frame-level and proposal-level, respectively. For vanilla attention, we use the classic non-local attention modules~\cite{DBLP:conf/cvpr/0004GGH18} to replace the f-PCM and p-PCM for comparison. The results are shown in Table~\ref{table:different_context_modeling_methods}. Our f-PCM outperforms the 1D non-local block by a large margin with comparable running speed. And our p-PCM achieves significant performance improvements with less additional time cost.

	\begin{table}[!t]
	\centering
	\caption{Validation of f-PCM and p-PCM in different frameworks on ActivityNet-1.3.}
	\resizebox{1\columnwidth}{!}{
		\begin{tabular}{l|ccc|c}
			\hline
			Method & 0.5 & 0.75 & 0.95 & Avg \\
			\hline
			DBG  			& 18.57 & 13.65 &  7.12 & 27.36\\	
			DBG+f-PCM 		& 18.84 & 13.79 &  7.23 & 28.18\\
			DBG+p-PCM 		& 19.00 & 13.87 &  7.31 & 28.25\\
			DBG+f-PCM+p-PCM & \textbf{19.24} & \textbf{13.98} & \textbf{7.39} & \textbf{28.53}\\		
			\hline	
			BSN  			& 21.54 & 15.17 & 7.16 & 30.03\\	
			BSN+f-PCM 		& 23.16 & 16.31 &  7.23  & 31.78\\
			BSN+p-PCM 		& 23.29 & 16.63 &  7.30  & 31.91\\
			BSN+f-PCM+p-PCM & \textbf{23.31} & \textbf{16.74} & \textbf{8.04} & \textbf{32.24}\\		
			\hline		
		\end{tabular}  
		
	}
	
	\label{different_framework}
\end{table}

\subsubsection{Generalization in Different Frameworks}
To further verify the generalization of our proposed position-sensitive context modeling module, we evaluate it with two more frameworks (i.e., DBG and BSN). As shown in Table~\ref{different_framework}, our proposed PCM modules can also works for these two frameworks and improves the average mAP of DBG and BSN by $1.17\%$ and $2.21\%$ on ActivityNet-1.3, respectively. 


\subsection{Qualitative Analysis}
Through the above experiments, we have proved the effectiveness of f-PCM and p-PCM. But what role does temporal position information play in the attention mechanism? We define extra experiments to show how position information help localize actions. 

To intuitively understand the behavior of p-PCM,
we visualize the average attention maps for all proposals in Figure~\ref{fig:vis_ppcm2}. Proposals that have the same start or/and end frame with the ground-truth action instance receive more attention in our p-PCM, which fits with the findings above, while the vanilla attention block integrates features dispersedly. These experiments demonstrate the proposed p-PCM module indeed has the ability to generate temporal-position-sensitive feature representation, and temporal position information is a vital cue for temporal action localization.
We also visualize the average attention maps of f-PCM modules before the start and end boundary classifier in Figure~\ref{fig:vis_fpcm}. The phenomenon is similar to p-PCM.

\section{Conclusion}
In this paper, we introduce the temporal position information which is a vital cue for more precise temporal action localization, and propose a simple yet effective position-sensitive context modeling approach to incorporates both positional and semantic context, in both frame-level and proposal-level. The extensive experiments and qualitative analysis demonstrate the importance of temporal position information and the effectiveness and intuitive working principle of our approach. We only use a very simple method to model the temporal information and use the position information to guide global information aggregation. How to better model and utilize the temporal position information need to be explored in future work.

\appendices
\section{Theoretical Analysis of Positional Encoding}
In this section, we analyze that the fixed sinusoid positional encoding can only offer the relative distance information without the direction theoretically. For simplicity, we take f-PCM module as an example to give a discussion.


When only considering the effect of positional encoding $\mathbf{pe}$ and computing the attention weight $f(\mathbf{\tilde{x}}_i,\mathbf{\tilde{x}}_j)$ (defined in line 460, Equation (5) of the main manuscript) between frame $v_i$ and frame $v_j$, we have: 
\begin{equation}
\begin{aligned}
f(\mathbf{pe}_i,\mathbf{pe}_j) 
& = e^{\phi(\mathbf{pe}_i) \theta(\mathbf{pe}_j)} \\
& = e^{ \mathbf{W}_{\phi}  \mathbf{pe}_i  \mathbf{pe}_j^{\mathsf{T}}  \mathbf{W}_{\theta}^{\mathsf{T}} }, \\
\end{aligned}
\label{fxpe}
\end{equation}
where $\mathbf{W}_{\phi}$ and $\mathbf{W}_{\theta}$ are the parameters of the embedding function $\phi$ and $\theta$, 
$\mathbf{pe}_i$ and $\mathbf{pe}_j$ are the fixed sinusoid positional encoding of frame $v_i$ and $v_j$, that use sine and cosine functions of different frequencies to encode the position information:
\begin{equation}
\left\{\begin{matrix}
pe_{i,2d}  &=\sin(\frac{i}{10000^{2d/l}})\\
pe_{i,2d+1} & =\cos(\frac{i}{10000^{2d/l}})
\end{matrix}\right.
, d=\{0,1,...,k-1\}
\end{equation}
where $2d$ and $2d+1$ represent the $2d$-th and $(2d+1)$-th dimension of the $l$-dimensional positional encoding $\mathbf{pe}_i$, respectively. 
We denote $b=10000$, $l=2k$ for simplicity, then we have:
\begin{equation}
\begin{aligned}
\mathbf{pe}_i  \mathbf{pe}_j^{\mathsf{T}} 
&= \frac{1}{\left \| \mathbf{pe}_i \right \|} \frac{1}{\left \| \mathbf{pe}_j \right \|} \cdot \\
&\sum_{d=0,...,k-1} \sin(\frac{i}{b^{d/k}}) \sin(\frac{j}{b^{d/k}})  + \cos(\frac{i}{b^{d/k}}) \cos(\frac{j}{b^{d/k}}) \\
& = \frac{1}{k} \sum_{d=0,1,...,k-1} \cos(\frac{i-j}{b^{d/k}})
\end{aligned}
\end{equation}
Since $\cos(\cdot)$ is an even function, we have:
\begin{equation}
\begin{aligned}
\mathbf{pe}_i  \mathbf{pe}_{i+r}^{\mathsf{T}} 
& = \frac{1}{k} \sum_{d=0,1,...,k-1} \cos(\frac{i-(i+r)}{b^{d/k}})  \\
& = \frac{1}{k} \sum_{d=0,1,...,k-1} \cos(\frac{i-(i-r)}{b^{d/k}}) \\
& = \mathbf{pe}_i  \mathbf{pe}_{i-r}^{\mathsf{T}}
\end{aligned}
\end{equation}
Substituting this result into Equation (\ref{fxpe}), we get:
\begin{equation}
\begin{aligned}
f(\mathbf{pe}_i,\mathbf{pe}_{i+r}) 
& = e^{ \mathbf{W}_{\phi}  \mathbf{pe}_i  \mathbf{pe}_{i+r}^{\mathsf{T}}  \mathbf{W}_{\theta}^{\mathsf{T}} } \\
& = e^{ \mathbf{W}_{\phi}  \mathbf{pe}_i  \mathbf{pe}_{i-r}^{\mathsf{T}}  \mathbf{W}_{\theta}^{\mathsf{T}} } \\
& = f(\mathbf{pe}_i,\mathbf{pe}_{i-r}) 
\end{aligned}
\end{equation}
For frame $v_i$, the attention weights calculated for frame $v_{i+r}$ and frame $v_{i-r}$ are the same. This means that as long as the distances are the same, regardless of the direction, the calculated attention weights are the same.


\ifCLASSOPTIONcaptionsoff
  \newpage
\fi



\bibliographystyle{IEEEtran}
\bibliography{IEEEabrv}
\end{document}